\documentclass[conference]{IEEEtran}
\IEEEoverridecommandlockouts

\usepackage{cite}
\usepackage{amsmath,amssymb,amsfonts}
\usepackage{algorithmic}
\usepackage{graphicx}
\usepackage{textcomp}
\usepackage{xcolor}
\def\BibTeX{{\rm B\kern-.05em{\sc i\kern-.025em b}\kern-.08em
    T\kern-.1667em\lower.7ex\hbox{E}\kern-.125emX}}
\begin{document}

\title{Learning temporal embeddings from electronic health records of chronic kidney disease patients\\
{\footnotesize \textsuperscript{}}
\thanks{This work is funded by German G-BA, grant number 01NVF21116. \protect\copyright~2026 IEEE. Personal use of this material is permitted. Permission from IEEE must be obtained for all other uses, in any current or future media, including reprinting or republishing this material for advertising or promotional purposes, creating new collective works, for resale or redistribution to servers or lists, or reuse of any copyrighted component of this work in other works.}
}

\author{\IEEEauthorblockN{1\textsuperscript{st} Aditya Kumar}
\IEEEauthorblockA{\textit{} \\
\textit{Hahn-Schickard}\\
Freiburg, Germany}
\\
\IEEEauthorblockA{\textit{Intelligent Embedded Systems Lab} \\
\textit{University of Freiburg}\\
Freiburg, Germany \\
}
\and
\IEEEauthorblockN{2\textsuperscript{nd} Mario A. Cypko}
\IEEEauthorblockA{\textit{} \\
\textit{Hahn-Schickard}\\
Freiburg, Germany}
\\
\IEEEauthorblockA{\textit{Intelligent Embedded Systems Lab} \\
\textit{University of Freiburg}\\
Freiburg, Germany \\
}
\and
\IEEEauthorblockN{3\textsuperscript{rd} Oliver Amft}
\IEEEauthorblockA{\textit{} \\
\textit{Hahn-Schickard}\\
Freiburg, Germany}
\\
\IEEEauthorblockA{\textit{Intelligent Embedded Systems Lab} \\
\textit{University of Freiburg}\\
Freiburg, Germany \\
}
}

\maketitle

\begin{abstract}
We investigate whether temporal embedding models trained on longitudinal electronic health records can learn clinically meaningful representations without compromising predictive performance, and how architectural choices affect embedding quality. Model-guided medicine requires representations that capture disease dynamics while remaining transparent and task-agnostic, whereas most clinical prediction models are optimised for a single task. Representation learning facilitates learning embeddings that generalise across downstream tasks, and recurrent architectures are well-suited for modelling temporal structure in observational clinical data. Using the MIMIC-IV dataset, we study patients with chronic kidney disease (CKD) and compare three recurrent architectures: a vanilla LSTM, an attention-augmented LSTM, and a time-aware LSTM (T-LSTM). All models are trained both as embedding models and as direct end-to-end predictors. Embedding quality is evaluated via CKD stage clustering and in-ICU mortality prediction. The T-LSTM produces more structured embeddings, achieving a lower Davies–Bouldin Index (DBI = 9.91) and higher CKD stage classification accuracy (0.74) than the vanilla LSTM (DBI = 15.85, accuracy = 0.63) and attention-augmented LSTM (DBI = 20.72, accuracy = 0.67). For in-ICU mortality prediction, embedding models consistently outperform end-to-end predictors, improving accuracy from 0.72–0.75 to 0.82–0.83, which indicates that learning embeddings as an intermediate step is more effective than direct end-to-end learning.
\end{abstract}


\section{Introduction}
Modern healthcare is increasingly shaped by the vision of model-guided medicine~(MGM)~\cite{b1,b2}. In MGM models move beyond static prediction to become continuous companions in patient care~\cite{b2,b3}. MGM also forms the foundation for digital patient models or digital twins, computational representations of individual patients that can predict disease trajectories, evaluate therapy options, and support personalised care~\cite{b4,b5,b6,b7,b8}. However, in most clinical applications models have been built for single, task-specific objectives e.g., monitoring or disease progression~\cite{b8, b9, b10}, mortality prediction~\cite{b11, b12}, readmission risk~\cite{b13}, etc. The task-specific training objectives demonstrate perform well, however, such learning is tailored to a particular task rather than to the broader dynamics of disease progression. Thus, the task-specific models are difficult to extend to new endpoints without retraining from scratch~\cite{b14}, which makes them unsuitable for dynamically changing clinical scenarios. Furthermore, because the internal representations of task-specific models are rarely examined~\cite{b15}, it remains unclear whether the task-specific models truly capture the underlying clinical state. 

Therefore, there is a need to move from task-specific modelling to representation-centric modelling. An embedding model is a type of model designed not merely to predict outcomes, but to learn dense vector representations~(i.e., embeddings) that encodes complex, high-dimensional input data~(e.g., longitudinal EHR) into a compact, continuous space~\cite{b16}. Recent work in representation learning in clinical applications have demonstrated the that one embedding can support multiple downstream tasks e.g., phenotyping~\cite{b17}. A few studies have explored the application of representation learning to clinical time series~\cite{b18}, however, systematic investigations remain limited.

EHRs represent one of the most prominent sources of temporal data in routine healthcare. EHRs capture a patient’s evolving physiological state through laboratory tests, vital signs, medications, procedures over time~\cite{b19, b20}. However, modelling real-world EHR data is challenging due to high dimensionality~\cite{b21}, irregular sampling~\cite{b21, b22, b23}, and complex temporal dynamics~\cite{b21}. Recurrent architectures, particularly Long Short-Term Memory~(LSTM) networks are widely used to model EHR data for predicting clinical outcomes from patient data sequences~\cite{b24}. Extensions of LSTMs including attention-augmented LSTMs~\cite{b25} and time-aware LSTMs~(T-LSTM)~\cite{b26} address challenges associated with long sequences and irregular sampling respectively. Nevertheless, LSTM-based modelling approaches remain predominantly task-specific, i.e., trained end-to-end for a single supervised objective and evaluated solely on predictive performance for the task~\cite{b27}. Consequently, LSTM-based approaches fall short of learning task-agnostic, clinically meaningful embeddings that can support the multi-objective reasoning required for MGM and digital patient modelling. 

\begin{figure}[h]
    \centering
    \includegraphics[width=\columnwidth]{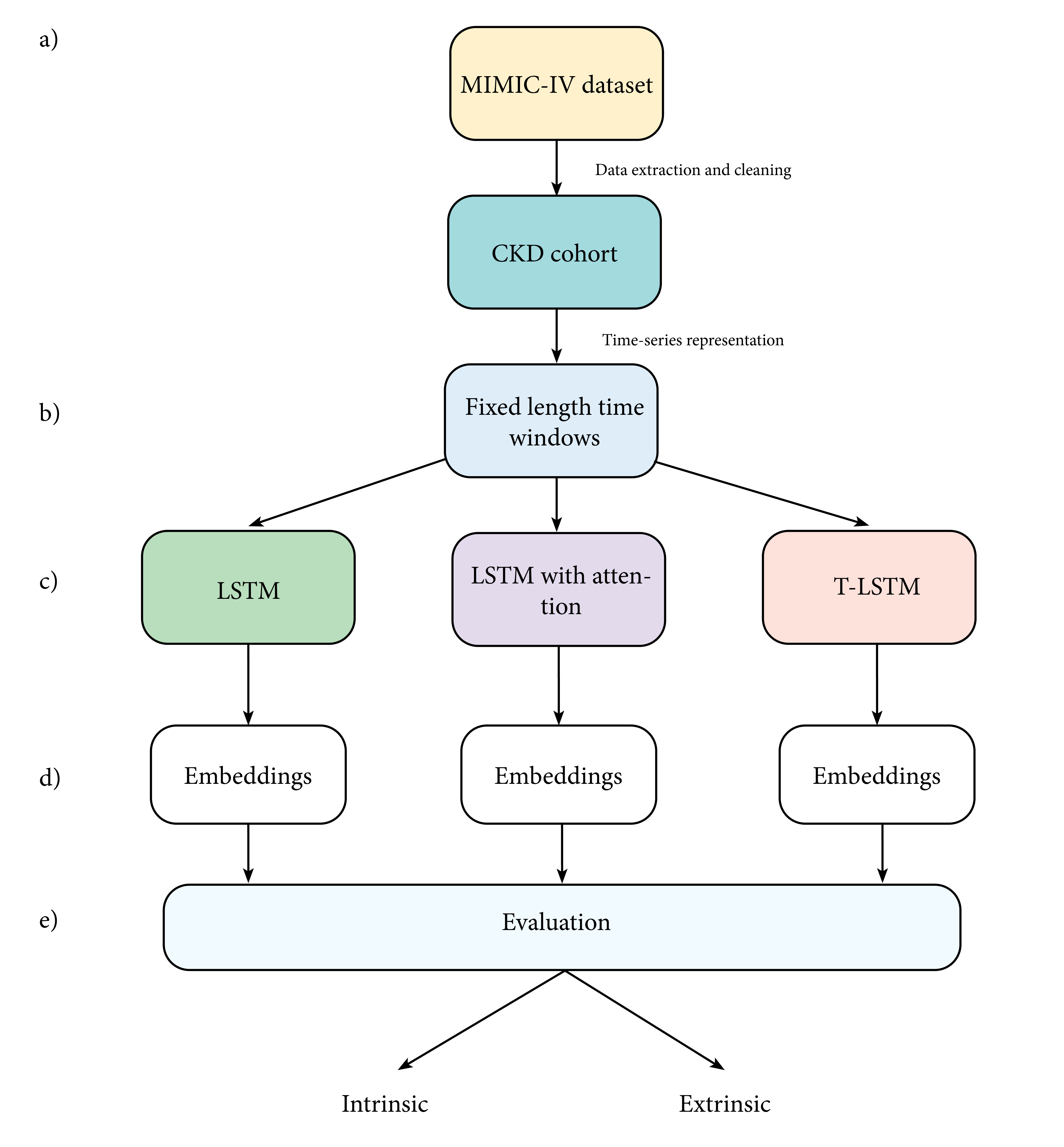}
    \caption{Overview of the approach. (a) Longitudinal EHR data are extracted and preprocessed to construct patient-specific time series. (b) Clinical trajectories are represented as irregularly sampled multivariate sequences with explicit time intervals. (c) Temporal models, including vanilla LSTM, attention-augmented LSTM, and T-LSTM, are trained as embedding models and end-to-end predictors. (d) The embedding models generate compact latent patient representations. (e) Embeddings are evaluated using intrinsic clustering analyses and extrinsic downstream prediction tasks.}
    \label{fig1}
\end{figure}

To address the aforementioned gaps, this work investigates whether embedding models trained on longitudinal EHR data can produce clinically meaningful representations without sacrificing predictive performance. We compared three recurrent architectures on a cohort of Chronic Kidney Disease~(CKD) patients: a vanilla LSTM, an attention-augmented LSTM, and a T-LSTM~(see Fig.~\ref{fig1}). 

All three models were trained in two configurations: as embedding models and as direct end-to-end predictor models. The embedding models were trained to learn patient representations by differentiating between CKD stages. The learnt representations were subsequently used for downstream prediction of in-ICU mortality. The end-to-end predictor models were trained directly to predict in-ICU mortality from longitudinal EHR data. We assessed the quality of learnt embeddings using two complementary evaluation strategies. First, clustering analysis to examine whether patients naturally grouped according to CKD stages within the learnt embedding space. Second, downstream prediction to test whether the embeddings retained sufficient information for in-ICU mortality prediction. The two-step evaluation separates the learning of patient representations from the outcome-specific classifier, to measure how well embeddings generalise beyond the training objective. 
Our work makes following contributions:
\begin{enumerate}
    \item We demonstrate that learning patient embeddings as an intermediate objective leads to better downstream prediction than direct end-to-end training across recurrent architectures.
    \item We demonstrate that explicit modelling of irregular time intervals~(T-LSTM) is particularly effective at learning structured and clinically meaningful representations from longitudinal EHR data.
\end{enumerate}

\section{Methods}
\subsection{Dataset}
We used MIMIC-IV v2.2~\cite{b28} for our study. We focused on a cohort of CKD patients. We utilised the pipeline of Gupta et al.~\cite{b29} to extract our cohort~(referred as CKD cohort). The CKD cohort consisted of 3932 patients diagnosed with CKD, amounting to 10,000 admission cases. The MIMIC-IV dataset is divided into two domains: hospital and ICU. The hospital domain consists of tables including patient’s personal information, records of hospital admission, records of ICD-9 and ICD-10 coded diagnoses, details on prescribed medicates, lab measurements. The ICU domain consist of tables including records of ICU stays, records of ICD-9 and ICD-10 coded diagnoses for each ICU stay, records of administered fluids and medications, chart events, procedure events. 

\subsection{Data processing} \label{dp}
Each patient admission in the CKD cohort was split into three structured files: demographic file~(consisting of demographic details specific to the patient including subject\_id, age, sex), static features~(consisting of all static features in one single hospital or ICU stay), and time-series features~(e.g., medications, lab events, procedures that change repeatedly over a stay). We converted the longitudinal data into fixed-length time-series representations. First, we defined a window of first 72 hours. The window defines the observation period for each patient’s admission. Then, we divided the selected time window into equally sized time buckets of 1 hour, which aggregated the raw events in a 72 hour window into 72 time intervals. Within each bucket, we aggregated event data~(using statistical operations including mean for continuous values and binary indicators for event occurrence) to form a compressed time-series representation. To ensure complete time-series representations, we applied a two-step imputation strategy consisting of forward filling followed by mean or median imputation based on patient-specific observations. A data generator then aligned dynamic clinical features with admission time, aggregated them into pre-defined time bins, applied imputation, and stored the resulting representations for each admission. At each time-step, dynamic clinical events were concatenated with static features and demographic data along the feature dimension to construct a vector per time-step. Furthermore, we defined two modes of operation: predicting and clustering. For CKD stages, the labels were derived from the ICD code. We unified the ICD-9 and ICD-10 codes~\cite{b30, b31} for CKD into eight classes~(0-7) to ensure consistency across the two ICD versions~(see Table \ref{tab1}). For in-ICU mortality, the labels were saved as binary~(0-1).

\begin{table}[h]
\centering
\caption{Mapping of CKD ICD-10 and ICD-9 codes to classes} 
\label{tab1}
\begin{tabular}{llll}
\hline
\textbf{ICD-10 code} & \textbf{ICD-9 code} & \textbf{CKD Stage} & \textbf{Class} \\
\hline
N18.1 & 585.1 & Stage 1 & 0 \\
N18.2 & 585.2 & Stage 2 & 1 \\
N18.3 & 585.3 & Stage 3 & 2 \\
N18.4 & 585.4 & Stage 4 & 3 \\
N18.5 & 585.5 & Stage 5 & 4 \\
N18.6 & 585.6 & End stage renal disease & 5 \\
N18.8 & -- & Other CKD & 6 \\
N18.9 & 585.9 & Unspecified & 7 \\
\hline
\end{tabular}
\end{table}

\subsection{Model architectures}
\begin{figure}[h]
    \centering
    \includegraphics[width=\columnwidth]{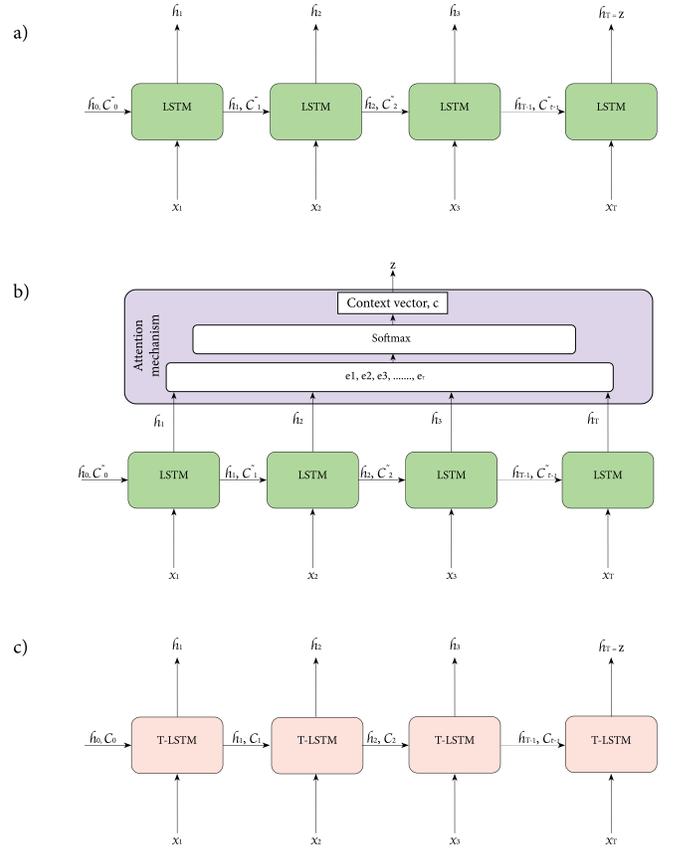}
    \caption{Recurrent model architectures evaluated in this study. (a) Vanilla LSTM, in which hidden and cell states are updated sequentially from the input time-series. (b) Attention-augmented LSTM, in which an attention mechanism over LSTM hidden states is applied to compute a context vector summarising the sequence. (c) Time-aware LSTM~(T-LSTM), in which time elapsed between observations are explicitly incorporated into the cell state update to model irregularly sampled clinical data.}
    \label{fig2}
\end{figure}

We used three recurrent architectures: a vanilla LSTM, an attention-augmented LSTM, and a T-LSTM. Fig.~\ref{fig2} illustrates the model architectures. 

The concatenated vector for each time-step was first passed through a fully connected layer to project the combined information into a fixed-size representation, and the resulting sequence of representations was used as input to the three model architectures. For the vanilla LSTM i.e., standard unidirectional LSTM, the model processes the input sequence timestep by timestep, effectively capturing temporal dependencies in the data~(see Fig.~\ref{fig2} a). Given an input sequence $\{x_1, x_2, \ldots, x_T\}$, the LSTM produces a corresponding sequence of cell states $\{\tilde{C}_1, \tilde{C}_2, \ldots, \tilde{C}_T\}$ and hidden states $\{h_1, h_2, \ldots, h_T\}$ through recurrent updates: $h_t=\mathrm{LSTM}(x_t, h_{t-1})$. We treated the final hidden state $h_T$, which summarises information from the entire sequence as the embedding representation, $z = h_T$. The embeddings captured the core temporal dynamics of our input data. The extracted embeddings were subsequently passed through two fully connected layers to generate a prediction vector: $u=\phi(W_1 z + b_1), \quad v = W_2 u + b_2 $, where $\phi(\cdot)$ denotes a non-linear activation function (e.g., ReLU). Finally, we applied a sigmoid activation function $\hat{y} = \sigma(v)$, to generate a probability value between 0 and 1 corresponding to the target outcome. 

The attention-augmented LSTM builds on the vanilla LSTM by adding an attention mechanism that facilitates the model to selectively weight different time steps according to their relevance to the final prediction. In addition to the basic steps, it adds: an attention layer, and a context vector computation~(see Fig.~\ref{fig2} b). After processing the input sequence, the LSTM produces a sequence of hidden states $\{h_1, h_2, \ldots, h_T\}$. For the attention layer, we added an additional linear layer after the LSTM layer to compute a scalar attention score for each hidden state: $e_t = w_a^\top h_t + b_a$.
The attention scores were normalised across the temporal dimension using a softmax function:
\begin{equation}
\alpha_t = \frac{\exp(e_t)}{\sum_{k=1}^{T} \exp(e_k)}.
\end{equation}
The normalised attention weights $\alpha_t$ reflect the relative importance of each time step. For the context vector computation, we used the normalised attention weights to compute a context vector via a weighted sum of all hidden states: $c = \sum_{t=1}^{T} \alpha_t h_t$. The context vector $c$ essesntially embeds the most informative aspects of the entire sequence by emphaising key time steps.

The T-LSTM extends the vanilla LSTM by explicitly incorporating time intervals between consecutive events into the LSTM cell computations~(see Fig.~\ref{fig2} c). In addition to the typical LSTM operations, the T-LSTM architecture introduces: a time decay mechanism, and a modified LSTM cell operation. The time decay mechanism is a learnable projection applied to the time interval $(\Delta t)$ between consecutive events: $\gamma_t = \exp\!\left(-\max\!\left(0, W_{\Delta} \Delta t_t + b_{\Delta}\right)\right)$. The projection produces a decay factor $\gamma_t \in (0,1]$, which is then used to adjust the previous cell state. The longer the time gap, the more the influence of the past information is reduced. Specifically, the previous cell state is decayed as:$C_{t-1} = \gamma_t \odot \tilde{C}_{t-1}$. The cell operation is adapted such that, before computing the input, forget, and output gates, the decayed cell state $C_{t-1}$ is used instead of the original cell state. The gate computations are therefore defined as:
\begin{equation}
f_t = \sigma(W_f x_t + U_f h_{t-1} + b_f),
\end{equation}
\begin{equation}
i_t = \sigma(W_i x_t + U_i h_{t-1} + b_i),
\end{equation}
\begin{equation}
o_t = \sigma(W_o x_t + U_o h_{t-1} + b_o).
\end{equation}
The updated cell and hidden states are then computed as:
\begin{equation}
C_t = f_t \odot \tilde{C}_{t-1} + i_t \odot \tanh(W_c x_t + U_c h_{t-1} + b_c),
\end{equation}
\begin{equation}
h_t = o_t \odot \tanh(c_t).
\end{equation}
The modified cell operation enables the model to naturally handle the irregular timing of clinical events.

\subsection{Training and embedding extraction}
During training, the three model architectures processed fixed-length time-series inputs to learn robust representations of patient trajectories. The data was partitioned into training, validation and test sets using 5-fold cross-validation. During each fold, the individual model was optimised using cross-entropy loss. An early stopping was also employed based on the validation loss to prevent overfitting. Moreover, if oversampling was activated, the minority class was upsmapled before training to counteract class imbalance. Once training was complete, the trained models were used to extract patient embeddings. For embedding extraction, the outputs of each individual model’s final layer were flattened and then the resulting embeddings were saved.

\subsection{Intrinsic evaluation of embeddings}
For intrinsic evaluation, we projected the high-dimensional embeddings onto a 2D space using t-SNE. The t-SNE visualisation facilitates the empirical observation to analyse whether the patients who have the same ICD code are clustered together, which gives us a qualitative measure of the embeddings’ representational quality. We then compared the embeddings learnt by the three models on our CKD cohort. We used the classes~(0-7) defined in Section~\ref{dp} for clustering and classification. Furthermore, we also report the Davies-Bouldin Index~(DBI)~\cite{b32} on the 2D t-SNE projections of the test-set embeddings and classification accuracy by training a logistic regression classifier. For the evaluation, each model was trained using 5-fold cross validation, then tested on held-out data.

\subsection{Extrinsic evaluation of embeddings}
For the extrinsic evaluation, we trained a logistic regression classifier using the generated embeddings as input features to predict in-ICU mortality within a 2-hour prediction window. The input data consisted of the first 72 hours of ICU records, aggregated into 1-hour time bins. Model performance was evaluated using accuracy, area under the receiver operating characteristic curve~(AUROC), and area under the precision–recall curve~(AUPRC). The performance metrics serve as an indirect assessment of the embeddings’ representational quality, reflecting how effectively they capture clinically relevant information. We compared the three models under two configurations: end-to-end predictor i.e., models directly made prediction, and embedding model i.e., a downstream logistic regression classifier was trained on the learnt embeddings.

\section{Results}

\subsection{Intrinsic evaluation}
 Table \ref{tab2} shows the DBI on t-SNE projections of the test-set embeddings and downstream accuracy by training a logistic regression classifier on the test-set embeddings.

\begin{table*}[h]
\centering
\caption{Davies-Bouldin Index~(DBI) and downstream accuracy on CKD stage classification across 5 folds for the three model architectures.}
\label{tab2}
\setlength{\tabcolsep}{6pt}
\renewcommand{\arraystretch}{1.15}
\begin{tabular}{c cc cc cc}
\hline
\textbf{Fold} &
\multicolumn{2}{c}{\textbf{LSTM}} &
\multicolumn{2}{c}{\textbf{Attention-augmented LSTM}} &
\multicolumn{2}{c}{\textbf{T-LSTM}} \\
\cline{2-7}
& \textbf{DBI} & \textbf{Accuracy} & \textbf{DBI} & \textbf{Accuracy} & \textbf{DBI} & \textbf{Accuracy} \\
\hline
0 & 51.37 & 0.59 & 5.57 & 0.68 & 17.90 & 0.70 \\
1 & 6.95  & 0.58 & 70.85 & 0.70 & 9.80  & 0.74 \\
2 & 10.88 & 0.64 & 9.93  & 0.68 & 14.02 & 0.72 \\
3 & 4.17  & 0.64 & 6.71  & 0.66 & 4.32  & 0.77 \\
4 & 5.89  & 0.71 & 10.55 & 0.66 & 3.51  & 0.80 \\
\hline
\textbf{Mean $\pm$ SD} &
\textbf{15.85 $\pm$ 20.01} & \textbf{0.63 $\pm$ 0.05} &
\textbf{20.72 $\pm$ 28.10} & \textbf{0.68 $\pm$ 0.02} &
\textbf{9.91 $\pm$ 6.18}  & \textbf{0.74 $\pm$ 0.04} \\
\hline
\end{tabular}
\end{table*}

The vanilla LSTM showed moderate clustering quality with an average DBI of 15.85 and achieved a downstream classification accuracy of 0.63. Adding attention mechanism on LSTM increased the average downstream classification accuracy to 0.67, which suggests better overall predictive power. The DBI varied consistently from fold to fold, including one extremely high value of 70.85~(fold 1) that raised the mean DBI to 20.72. Incorporating the temporal gaps between events substantially enhanced both cluster separation~(mean DBI of 9.91) and classification accuracy ~0.74). Notably, fold 4 achieved the best single-fold DBI of 3.51 and the highest accuracy 0.80.

Fig.~\ref{fig2} illustrates the t-SNE visualisations of the embedding spaces for the three model architectures. We observed that certain classes including CKD stage 1~(class 0) and unspecified CKD~(class 7) remained dispersed in the t-SNE space. Overall, the clusters showed that the T-LSTM embedding space~(Fig.~\ref{fig2} c) had the best structure~(i.e., well-separated classes), while vanilla LSTM~(Fig.~\ref{fig2} a) showed relatively poor structure in the embedding space.

\begin{figure*}[h]
    \centering
    \includegraphics[width=\textwidth]{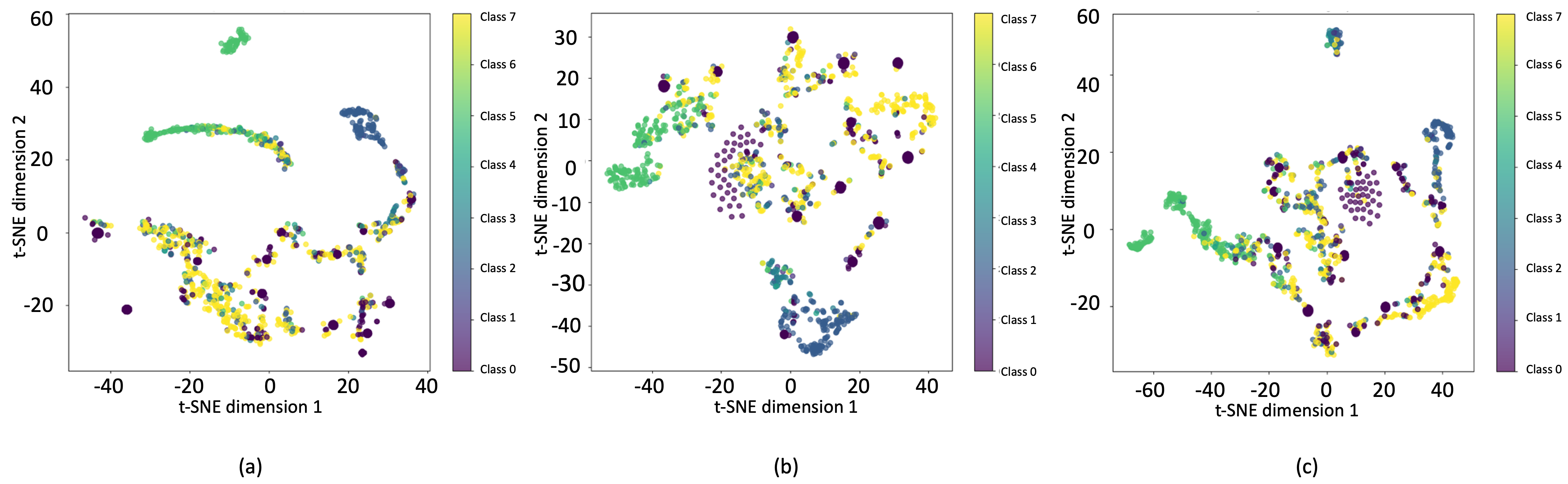}
    \caption{t-SNE visualisation of embedding spaces separated by classes 0-7 based on ICD codes for CKD stages. (a) Embedding space of vanilla LSTM. (b) Embedding space of attention-augmented LSTM. (c) Embedding space of T-LSTM.}
    \label{fig3}
\end{figure*}

\subsection{Extrinsic evaluation}
Table~\ref{tab3} summarises the 5-fold average performance of the three model architectures as an embedding model with downstream logistic regression, and as a direct end-to-end predictor for in-ICU mortality prediction. 

\begin{table*}[t]
\centering
\caption{Comparison of in-ICU mortality prediction across the three model architectures~(5-fold average). Direct prediction refers to models trained in a direct end-to-end predictor configuration, while downstream LR refers to models trained as embedding models with a logistic regression prediction head.}
\label{tab3}
\setlength{\tabcolsep}{8pt}
\renewcommand{\arraystretch}{1.15}
\begin{tabular}{l ccc ccc}
\hline
\textbf{Model} &
\multicolumn{3}{c}{\textbf{Direct prediction}} &
\multicolumn{3}{c}{\textbf{Downstream LR}} \\
\cline{2-7}
& \textbf{AUROC} & \textbf{Accuracy} & \textbf{AUPRC}
& \textbf{AUROC} & \textbf{Accuracy} & \textbf{AUPRC} \\
\hline
LSTM & 0.88 & 0.74 & \textbf{0.83} & 0.89 & \textbf{0.83} & 0.89 \\
Attention-augmented LSTM & \textbf{0.89} & 0.72 & 0.82 & \textbf{0.90} & 0.82 & 0.89 \\
T-LSTM & 0.88 & \textbf{0.75} & 0.82 & \textbf{0.90} & 0.82 & \textbf{0.90} \\
\hline
\end{tabular}
\end{table*}

In end-to-end predictor configuration, all models achieved comparable AUROC values~(0.88-0.89), with the T-LSTM yielding the highest accuracy~(0.75) compared to the vanilla LSTM~(0.74) and attention-augmented LSTM~(0.72). In the embedding model configuration with a downstream logistic regression head, the performance consistently improved across all three architectures. AUROC increased to 0.89 for the vanilla LSTM and to 0.90 for both the attention-augmented LSTM and T-LSTM. Accuracy improved from 0.72–0.75 to 0.82–0.83. Similarly, AUPRC values increased from 0.82–0.83 in the end-to-end predictor configuration to 0.89–0.90 in the embedding model with logistic regression configuration. Overall, the findings indicate that learning patient embeddings as an intermediate objective leads to better downstream prediction than direct end-to-end training across all three architectures. Furthermore, the T-LSTM achieved the strongest and most consistent performance across evaluation metrics.

\section{Discussion}
The intrinsic evaluation indicates that the T-LSTM model produces the most coherent embedding space with respect to CKD stage differentiation, which highlights the importance of explicitly modelling irregular temporal intervals in longitudinal clinical data. Among the three architectures, T-LSTM achieved the highest CKD stage classification accuracy~(0.74) and the lowest Davies–Bouldin Index~(DBI; 9.91 compared to 20.72 for attention-augmented LSTM and 15.85 for vanilla LSTM), which suggests improved cluster compactness and separability. 

The findings show that incorporating temporal information directly into the recurrent dynamics leads to more structured and clinically meaningful embeddings. The attention-augmented outperformed the vanilla LSTM in terms of classification accuracy~(0.67 vs. 0.63), however, its DBI exhibited greater variability across folds. The variability suggests that while attention mechanisms enhanced discriminative performance, they may also introduce instability in the learnt embedding structure, potentially due to over-emphasising specific temporal segments~\cite{b33}. Visual inspection of the t-SNE projections further revealed that early-stage and unspecified CKD cases were more diffusely distributed in the embedding space, a pattern that is clinically plausible given the inherent diagnostic ambiguity associated with these categories. In contrast, well-defined CKD stages formed tighter clusters, which underscores the influence of diagnostic clarity on representation learning in clinical settings.

In the extrinsic evaluation, the embedding model configuration coupled with a downstream logistic regression classifier consistently outperformed the corresponding end-to-end prediction models across all three architectures in the in-ICU mortality prediction task. The result demonstrates that extracting and classifying on learnt embedding can yield superior performance, likely because of the induced implicit knowledge of CKD stages allows the downstream classifier to better exploit the latent features. Importantly, the finding also provides indirect evidence of the quality of the learnt embeddings. In the in-ICU mortality setting, the performance of T-LSTM and attention-augmented LSTM was comparable, which may be attributed to the relatively unambiguous nature of the mortality outcome compared to CKD stage classification. Nevertheless, T-LSTM exhibited more robust performance across both intrinsic and extrinsic tasks, suggesting greater generalisability across heterogeneous clinical objectives.

\subsection{Limitations}
This study has some limitations that warrant consideration. First, the models were evaluated on a single cohort, which may limit generalisability of the learnt embeddings to other clinical populations, institutions, or data protocols. Second, while intrinsic evaluations using clustering metrics and visualisations provide useful insights into latent space organisation, they remain indirect proxies for clinical relevant and may not fully capture all clinically meaningful factors~\cite{b34}. Third, the embedding models were trained in a supervised manner~(i.e., learning to differentiate between CKD stages), and therefore the resulting representations are not strictly task-agnostic. Certain latent dimensions may encode label-specific information rather than purely generalisable patterns. Fourth, missing data were addressed through pre-processing and implicit modelling assumption rather than being explicitly treated as an informative signal, despite the fact that missingness often carry clinical meaning in real-world healthcare settings~\cite{b35}. Fifth, the increased architectural complexity introduced by attention mechanisms and time-aware components may also increase sensitivity to hyperparameter choices and pose challenges for interpretability. And finally, although predictive performance and embedding quality were systematically evaluated, prospective validation and structured assessment by clinical experts were beyond the scope of this study and are necessary step toward showing clinical utility.

\subsection{Future work}
Truly task-agnostic representations would require unsupervised or self-supervised learning objectives~\cite{b36}. Future work will therefore focus on developing label-free embedding models that rely solely on intrinsic temporal patterns in the data. In addition, clinical data are inherently multi-modal, encompassing not only longitudinal physiological measurements but also clinical text and narrative information. Extending the proposed framework to integrate multiple data modalities~\cite{b37} represents a natural and important next step. Beyond predictive performance and clustering, understanding how embeddings are structured and whether they capture meaningful, complete, and interpretable clinical factors remains an open question. Disentanglement-based representation learning~\cite{b38, b39} offers a promising avenue for addressing the challenge and will be explored in future studies. Furthermore, model interpretability is essential for clinical decision support, accordingly, future work will investigate feature attribution methods and compare model-derived importance scores with clinician assessments to evaluate clinical plausibility. Finally, assessing and improving model calibration~\cite{b40} will be a key focus to ensure reliable probability estimates in real-world clinical deployment.


\section*{Conclusion}
In this study, we examined whether temporal embedding models can generate clinically meaningful representations from longitudinal electronic health records without sacrificing predictive performance. By systematically comparing vanilla LSTM, attention-augmented LSTM, and T-LSTM architectures on CKD patients from the MIMIC-IV database, we demonstrated that explicitly modelling temporal intervals leads to more structured and discriminative embedding spaces. For both intrinsic clustering analyses and extrinsic in-ICU mortality prediction, T-LSTM consistently produced clearer CKD stage separation and better predictive performance compared to attention-augmented LSTM and vanilla LSTM architectures. Furthermore, embedding model configurations outperformed their end-to-end predictor counterparts.  Together, the findings suggest that temporal embedding learning offers a foundation for patient representations that support downstream tasks and provide a scalable basis for digital patient models, a step forward in MGM.


\begin{thebibliography}{00}
\bibitem{b1} Lemke, H.U. and Golubnitschaja, O., 2014. Towards personal health care with model-guided medicine: long-term PPPM-related strategies and realisation opportunities within ‘Horizon 2020’. EPMA Journal, 5(1), p.8.
\bibitem{b2} Cypko, M.A. and Wilhelm, D., 2024. Ladies and Gentlemen! This is no humbug. Why Model-Guided Medicine will become a main pillar for the future healthcare system. International Journal of Computer Assisted Radiology and Surgery, 19(10), pp.1919-1927.
\bibitem{b3} Chase, J.G., Preiser, J.C., Dickson, J.L., Pironet, A., Chiew, Y.S., Pretty, C.G., Shaw, G.M., Benyo, B., Moeller, K., Safaei, S. and Tawhai, M., 2018. Next-generation, personalised, model-based critical care medicine: a state-of-the art review of in silico virtual patient models, methods, and cohorts, and how to validation them. Biomedical engineering online, 17(1), p.24.
\bibitem{b4} Vallée, A., 2024. Envisioning the future of personalized medicine: role and realities of digital twins. Journal of medical Internet research, 26, p.e50204.
\bibitem{b5} Subasi, A. and Subasi, M.E., 2024. Digital twins in healthcare and biomedicine. In Artificial Intelligence, Big Data, Blockchain and 5G for the Digital Transformation of the Healthcare Industry (pp. 365-401). Academic Press.
\bibitem{b6} Li, X., Loscalzo, J., Mahmud, A.F., Aly, D.M., Rzhetsky, A., Zitnik, M. and Benson, M., 2025. Digital twins as global learning health and disease models for preventive and personalized medicine. Genome Medicine, 17(1), p.11.
\bibitem{b7} Sharma, H. and Kaur, S., 2025. Patient-specific digital twins for personalized healthcare: a hybrid AI and simulation-based framework. IEEE Access.
\bibitem{b8} Iyer, A.A. and Umadevi, K.S., 2025. Design and analysis of TwinCardio framework to detect and monitor cardiovascular diseases using digital twin and deep neural network. Scientific Reports, 15(1), p.24376.
\bibitem{b9} Li, Y., Rao, S., Solares, J.R.A., Hassaine, A., Ramakrishnan, R., Canoy, D., Zhu, Y., Rahimi, K. and Salimi-Khorshidi, G., 2020. BEHRT: transformer for electronic health records. Scientific reports, 10(1), p.7155.
\bibitem{b10} Ma, F., Chitta, R., Zhou, J., You, Q., Sun, T. and Gao, J., 2017, August. Dipole: Diagnosis prediction in healthcare via attention-based bidirectional recurrent neural networks. In Proceedings of the 23rd ACM SIGKDD international conference on knowledge discovery and data mining (pp. 1903-1911).
\bibitem{b11} Zhang, D., Yin, C., Zeng, J., Yuan, X. and Zhang, P., 2020. Combining structured and unstructured data for predictive models: a deep learning approach. BMC medical informatics and decision making, 20(1), p.280.
\bibitem{b12}Si, Y., Du, J., Li, Z., Jiang, X., Miller, T., Wang, F., Zheng, W.J. and Roberts, K., 2021. Deep representation learning of patient data from Electronic Health Records (EHR): A systematic review. Journal of biomedical informatics, 115, p.103671.
\bibitem{b13} Deznabi, I., Iyyer, M. and Fiterau, M., 2021, August. Predicting in-hospital mortality by combining clinical notes with time-series data. In Findings of the association for computational linguistics: ACL-IJCNLP 2021 (pp. 4026-4031).
\bibitem{b14} Shivashankar, K., Hajj, G.S.A. and Martini, A., 2025. Scalability and Maintainability Challenges and Solutions in Machine Learning: Systematic Literature Review. arXiv preprint arXiv:2504.11079.
\bibitem{b15} Niv, Y., 2019. Learning task-state representations. Nature neuroscience, 22(10), pp.1544-1553.
\bibitem{b16} Kauffman, J., Miotto, R., Klang, E., Costa, A., Norgeot, B., Zitnik, M., Khader, S., Wang, F., Nadkarni, G.N. and Glicksberg, B.S., 2025. Embedding Methods for Electronic Health Record Research. Annual Review of Biomedical Data Science, 8.
\bibitem{b17} Zheng, X., 2025. Machine Learning for Complex Clinical Time Series: From Representation Learning to Interpretable Phenotyping.
\bibitem{b18} Ruan, T., Lei, L., Zhou, Y., Zhai, J., Zhang, L., He, P. and Gao, J., 2019. Representation learning for clinical time series prediction tasks in electronic health records. BMC medical informatics and decision making, 19(Suppl 8), p.259.
\bibitem{b19} Xie, F., Yuan, H., Ning, Y., Ong, M.E.H., Feng, M., Hsu, W., Chakraborty, B. and Liu, N., 2022. Deep learning for temporal data representation in electronic health records: A systematic review of challenges and methodologies. Journal of biomedical informatics, 126, p.103980.
\bibitem{b20} [16] Lee, S.A., Jain, S., Chen, A., Ono, K., Biswas, A., Rudas, Á., Fang, J. and Chiang, J.N., 2025. Clinical decision support using pseudo-notes from multiple streams of EHR data. npj Digital Medicine, 8(1), p.394.
\bibitem{b21} Shen, Y., Yu, J., Zhou, J. and Hu, G., 2025. Twenty-five years of evolution and hurdles in electronic health records and interoperability in medical research: comprehensive review. Journal of Medical Internet Research, 27, p.e59024.
\bibitem{b22} Zheng, Z., Luo, J., Zhu, Y., Du, L., Lan, L., Zhou, X., Yang, X. and Huang, S., 2025. Development and Validation of a Dynamic Real-Time Risk Prediction Model for Intensive Care Units Patients Based on Longitudinal Irregular Data: Multicenter Retrospective Study. Journal of medical Internet research, 27, p.e69293.
\bibitem{b23} Zhong, S., Wang, L.R., Zhan, Z., Ng, Y.Y. and Fan, X., 2025. A Hybrid Approach for Irregular-Time Series Prediction using Electronic Health Records: an Intensive Care Unit Mortality Case Study. ACM Transactions on Computing for Healthcare, 6(4), pp.1-33.
\bibitem{b24} Ren, W., Zhu, J., Liu, Z., Zhao, T. and Honavar, V., 2025. A comprehensive survey of electronic health record modeling: From deep learning approaches to large language models. arXiv preprint arXiv:2507.12774.
\bibitem{b25} Islam, M.S., Umran, H.M., Umran, S.M. and Karim, M., 2019, May. Intelligent healthcare platform: cardiovascular disease risk factors prediction using attention module based LSTM. In 2019 2nd international conference on artificial intelligence and big data (ICAIBD) (pp. 167-175). IEEE.
\bibitem{b26} Baytas, I.M., Xiao, C., Zhang, X., Wang, F., Jain, A.K. and Zhou, J., 2017, August. Patient subtyping via time-aware LSTM networks. In Proceedings of the 23rd ACM SIGKDD international conference on knowledge discovery and data mining (pp. 65-74).
\bibitem{b27} Zheng, Y., Bensahla, A., Bjelogrlic, M., Zaghir, J., Turbe, H., Bednarczyk, L., Gaudet-Blavignac, C., Ehrsam, J., Marchand-Maillet, S. and Lovis, C., 2025. A scoping review of self-supervised representation learning for clinical decision making using EHR categorical data. NPJ Digital Medicine, 8(1), pp.362-362.
\bibitem{b28} Johnson, A.E., Bulgarelli, L., Shen, L., Gayles, A., Shammout, A., Horng, S., Pollard, T.J., Hao, S., Moody, B., Gow, B. and Lehman, L.W.H., 2023. MIMIC-IV, a freely accessible electronic health record dataset. Scientific data, 10(1), p.1.
\bibitem{b29} Gupta, M., Gallamoza, B., Cutrona, N., Dhakal, P., Poulain, R. and Beheshti, R., 2022, November. An extensive data processing pipeline for mimic-iv. In Machine learning for health(pp. 311-325). PMLR.
\bibitem{b30} Stephens, J.H., Ledlow, G.R. and Fockler, T.V., 2016. Converting ICD-9 to ICD-10. Hospital topics, 94(1), pp.1-7.
\bibitem{b31} Zaiss, A., Schulz, S., Graubner, B. and Klar, R., 1996. Conversion Table between ICD-9 and ICD-10. In Medical Informatics Europe’96 (pp. 193-197). IOS Press.
\bibitem{b32} Chicco, D., Campagner, A., Spagnolo, A., Ciucci, D. and Jurman, G., 2025. The Silhouette coefficient and the Davies-Bouldin index are more informative than Dunn index, Calinski-Harabasz index, Shannon entropy, and Gap statistic for unsupervised clustering internal evaluation of two convex clusters. PeerJ Computer Science, 11, p.e3309.
\bibitem{b33} Kim, J., Lee, M. and Heo, J.P., 2023. Self-feedback detr for temporal action detection. In Proceedings of the IEEE/CVF International Conference on Computer Vision (pp. 10286-10296).
\bibitem{b35} Sim, T., Hahn, S., Kim, K.J., Cho, E.Y., Jeong, Y., Kim, J.H., Ha, E.Y., Kim, I.C., Park, S.H., Cho, C.H. and Yu, G.I., 2025. Preserving informative presence: How missing data and imputation strategies affect the performance of an AI-based early warning score. Journal of Clinical Medicine, 14(7), p.2213.
\bibitem{b34} Hernandez, B., Stiff, O., Ming, D.K., Ho Quang, C., Nguyen Lam, V., Nguyen Minh, T., Nguyen Van Vinh, C., Nguyen Minh, N., Nguyen Quang, H., Phung Khanh, L. and Dong Thi Hoai, T., 2023. Learning meaningful latent space representations for patient risk stratification: Model development and validation for dengue and other acute febrile illness. Frontiers in Digital Health, 5, p.1057467.
\bibitem{b36} Ericsson, L., Gouk, H., Loy, C.C. and Hospedales, T.M., 2022. Self-supervised representation learning: Introduction, advances, and challenges. IEEE Signal Processing Magazine, 39(3), pp.42-62.
\bibitem{b37} Moor, M., Banerjee, O., Abad, Z.S.H., Krumholz, H.M., Leskovec, J., Topol, E.J. and Rajpurkar, P., 2023. Foundation models for generalist medical artificial intelligence. Nature, 616(7956), pp.259-265.
\bibitem{b38} Zheng, C., Zhu, Q., Fei, L., Li, S., Zhai, X.B., Zhang, D. and Zhang, D., 2025. Disentangled Representation Learning for Robust Brainprint Recognition. IEEE Transactions on Information Forensics and Security.
\bibitem{b39} Chen, H., Sun, J., Liu, Y., Shi, H. and Hu, D., 2025. Self-supervised disentangled representation learning via compositional invariance. IEEE Transactions on Circuits and Systems for Video Technology.
\bibitem{b40} Van Calster, B., McLernon, D.J., Van Smeden, M., Wynants, L. and Steyerberg, E.W., 2019. Calibration: the Achilles heel of predictive analytics. BMC medicine, 17(1), p.230.


\end{thebibliography}
\end{document}